\documentclass{article} 
\usepackage{arxiv,times}

\usepackage{amsmath,amsfonts,bm}

\def\eqref#1{equation~\ref{#1}}

\def\1{\bm{1}}

\DeclareMathAlphabet{\mathsfit}{\encodingdefault}{\sfdefault}{m}{sl}
\SetMathAlphabet{\mathsfit}{bold}{\encodingdefault}{\sfdefault}{bx}{n}

\usepackage[utf8]{inputenc} 
\usepackage[T1]{fontenc}    
\usepackage{hyperref}       
\usepackage{url}            
\usepackage{booktabs}       
\usepackage{amsfonts}       
\usepackage{nicefrac}       
\usepackage{microtype}      
\usepackage{xcolor}         
\usepackage{graphicx}
\usepackage{amsmath}
\usepackage{multirow}
\usepackage{amssymb}
\usepackage{float}
\usepackage{caption}
\usepackage{subcaption}
\usepackage[symbol]{footmisc}

\title{Multi-Tailed, Multi-Headed, Spatial Dynamic Memory refined Text-to-Image Synthesis}

\author{Amrit Diggavi Seshadri , Balaraman Ravindran \\
\texttt{asd20@ic.ac.uk, ravi@cse.iitm.ac.in} \\
Robert Bosch Center for Data Science and Artificial Intelligence,\\
Indian Institute of Technology Madras
}

\iclrfinalcopy 
\begin{document}

\maketitle

\begin{abstract}
Synthesizing high-quality, realistic images from text-descriptions is a challenging task, and current methods synthesize images from text in a multi-stage manner, typically by first generating a rough initial image and then refining image details at subsequent stages. However, existing methods that follow this paradigm suffer from three important limitations. Firstly, they synthesize initial images without attempting to separate image attributes at a word-level. As a result, object attributes of initial images (that provide a basis for subsequent refinement) are inherently entangled and ambiguous in nature. Secondly, by using common text-representations for all regions, current methods prevent us from interpreting text in fundamentally different ways at different parts of an image. Different image regions are therefore only allowed to assimilate the same type of information from text at each refinement stage. Finally, current methods generate refinement features only once at each refinement stage and attempt to address all image aspects in a single shot. This single-shot refinement limits the precision with which each refinement stage can learn to improve the prior image. Our proposed method introduces three novel components to address these shortcomings: (1) An initial generation stage that explicitly generates separate sets of image features for each word n-gram. (2) A spatial dynamic memory module for refinement of images. (3) An iterative multi-headed mechanism to make it easier to improve upon multiple image aspects. Experimental results demonstrate that our Multi-Headed Spatial Dynamic Memory image refinement with our Multi-Tailed Word-level Initial Generation (MSMT-GAN) performs favourably against the previous state of the art on the CUB and COCO datasets. 
\end{abstract}

\section{Introduction}
Generative Adversarial Networks (GANs) have shown great promise for the generation of photo-realistic synthetic images ~\citep{goodfellow2014generative, radford2015unsupervised, denton2015deep, salimans2016improved}, and the highly-compelling nature of images generated by GANs has driven research into conditional image-generation and multimodal learning. In this paper, we focus on the task of text-to-image generation, that has emerged as an area of active research in recent years. Although much progress has been made in this area, the synthesis of high-quality, realistic images from text-descriptions remains a challenging task.
Current state-of-the-art methods \citep{xu2018attngan, li2019controllable, zhu2019dm} employ multiple-stages of image generation - typically, an initial image is first generated from a global sentence-level vector, and subsequent stages incorporate fine-grained information extracted from word-level vectors to refine image details. However, these methods suffer from three important limitations. The first problem is that by attempting to synthesize image features directly from a sentence-level vector, the initial generation stage fails to cleanly separate image attributes at a word-level. If potentially distinct objects such as `cat' and `tree' for example, are entangled in the sentence-level representation, then the presence of either word in a sentence could prompt the initial stage to generate the same hybrid image attributes. This is important because the subsequent refinement stage relies upon the initial image features to provide a meaningful basis for word-level refinement. By feeding it ambiguous and poorly formed initial features, we limit the scope of refinement. Secondly, current methods do not construct region-specific representations of text at refinement stages. This prevents us from interpreting words in fundamentally different ways based on the content of image regions. Whereas, in complex real-world scenes, the requirement for a region-contextualized interpretation of words is commonplace - based on the region under consideration, the same word often dictates fundamentally different types of refinement within a single image. The word `raining' for example dictates a requirement in the sky that is fundamentally different from the requirement that it dictates in the region of the ground. While the sky becomes more cloudy, the ground must become wet. To generate realistic images from natural text descriptions, it is important that we construct a refinement architecture that allows different image regions to assimilate region-contextualized information from text descriptions. Finally, we note that current methods generate refinement features (that modify previous image features) only once at each refinement stage and attempt to address all image aspects within a single-shot. This single-shot refinement limits the precision with which each refinement stage can learn to improve the prior image.
\\\\
In this paper, we propose a Multi-Headed and Spatial Dynamic Memory image refinement mechanism with a Multi-Tailed Word-level Initial Generation stage (MSMT-GAN) to address these three issues. Our contributions are summarized as follows:
\begin{itemize}
    \item We introduce a novel "Multi-Tailed" Word-level Initial Generation stage (MTWIG), that generates a separate set of image features for each word n-gram, and iteratively fuses these sets together to obtain initial image features. We demonstrate that it is possible to improve the performance of previous methods by replacing their initial generation stage with ours.
    \item  We introduce a novel Spatial Dynamic Memory module (SDM) that fuses word-information in a custom way with each prior image region, to obtain region-contextualized text-representations. At each refinement stage we retrieve features for image improvement from this SDM module.
    \item We introduce a novel Iterative Multi-Headed Mechanism (IMHM) of image refinement - wherein we explicitly allow each stage of refinement to make multiple distinct modifications to the prior image, under common discriminator feedback.
\end{itemize}
We evaluate our MSMT-GAN model on the Caltech-UCSD Birds 200 (CUB) dataset \citep{wah2011caltech} and the Microsoft Common Objects in Context (COCO) dataset \citep{lin2014microsoft}. Experiment results demonstrate that MSMT-GAN is competitive with current methods on the COCO dataset and significantly outperforms the previous state-of-the art on the CUB dataset, decreasing the lowest reported Fréchet Inception Distance (FID) \citep{heusel2017gans} by $21.58\%$ for CUB.  

\section{Related Work}
\textbf{Text-to-Image Generators:}\cite{reed2016generative}  first demonstrated that a translation model from natural language to image pixels could be learnt by conditioning both generator and discriminator networks of a GAN on input text-descriptions. There has since been a surge of interest in training multi-stage attention based GAN architectures for this task. While the conventional setting \citep{zhang2017stackgan, xu2018attngan, li2019controllable, zhu2019dm} assumes only the availability of (text,image) pairs at training time, recently a second setting has emerged that assumes availability of bounding-box/shape-mask information of objects attributes during training \citep{li2019object, hinz2019semantic, cho2020x, liang2020cpgan}. We highlight that this represents a significantly easier problem setting and that such methods are not feasible where bounding-box/shape information is unavailable (such as the CUB dataset). Our method does not assume the availability of bounding-box/shape information, and we make comparisons against prior work of the same setting.\\\\
\textbf{Memory Networks:} Memory Networks \citep{weston2014memory} combine inference components with a long-term memory module that can be dynamically written to and read from. Current methods \citep{miller2016key} query ``key encodings" of memory slots to retrieve a set of weights. These weights are used to combine separate ``value encodings" of the slots into a single response.  A Dynamic Memory Generative Adversarial Network (DM-GAN) \citep{zhu2019dm} that retrieves information for image refinement from a memory  module was recently proposed for text-to-image synthesis. In our SDM module, we too employ the \textit{memory-writing, key-addressing, value-reading} paradigm introduced by \citep{miller2016key}, but our method differs from \citep{zhu2019dm} in all three memory operations (Section \ref{subsection:SDM}). Fundamentally, DM-GAN does not create region-contextualized representations of text.\\\\
\textbf{Multi-Headed Attention:} Transformers \citep{vaswani2017attention} utilize a key-value mechanism similar to memory networks and introduced the idea of multi-headed attention. They linearly project query, keys and values to $h$ separate encodings, called ``attention heads", and each head is separately used to extract an output vector. These vectors are concatenated together and linearly projected to a single response. Inspired by the success of Transformers, we introduce the IMHM method for image refinement. However, our method differs in a few respects. We maintain separate SDM modules for each head and we obtain queries and fuse outputs in an iterative fashion. We also adopt a ``redundancy loss" (Section \ref{subsection:OBJfunc}) to encourage each head to focus on separate image aspects.
\section{MSMT-GAN}
\label{section:MSDM-GAN-label}
\begin{figure*}
    \centering
    \includegraphics[width=\linewidth]{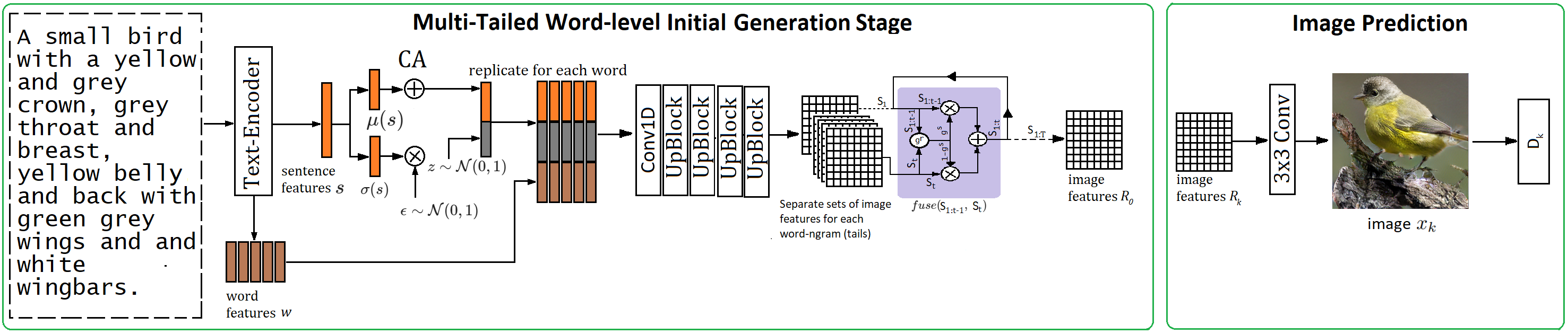}\\
    \includegraphics[width=\linewidth]{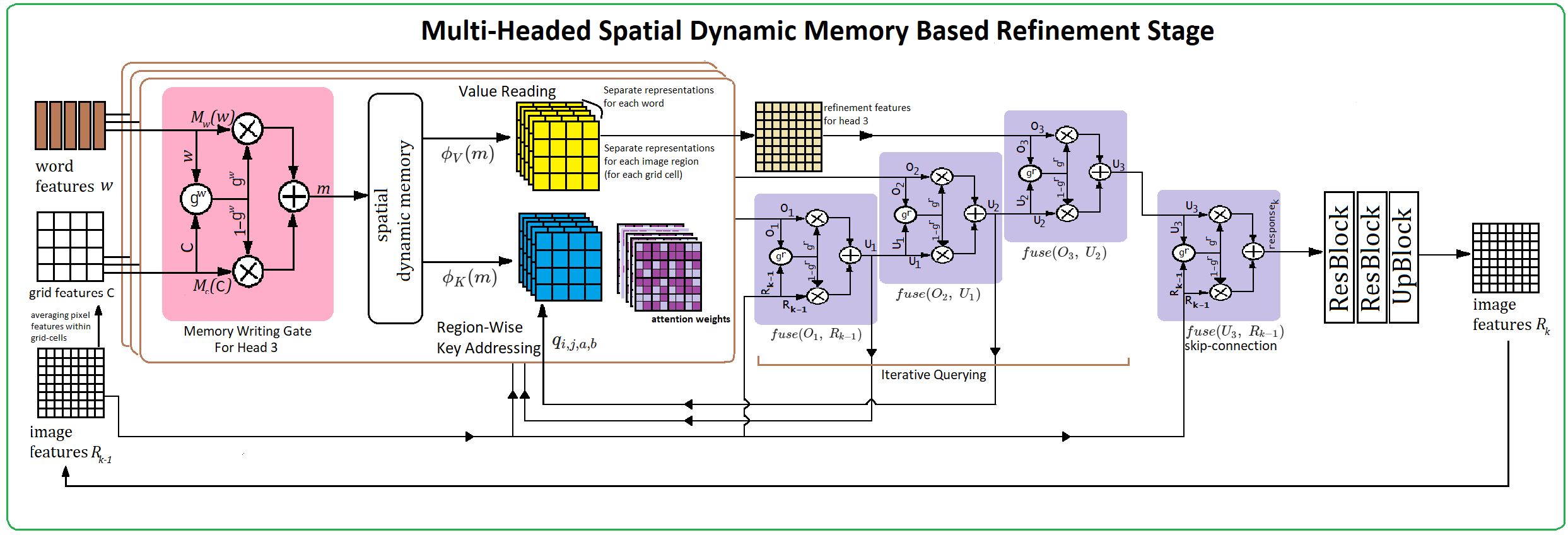}\\
    \includegraphics[width=\linewidth]{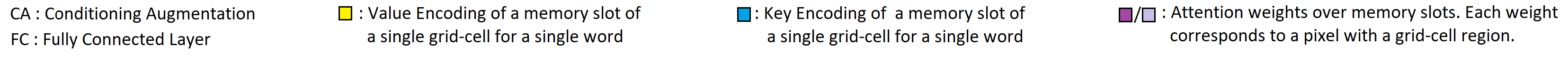}
    \caption{Our MSMT-GAN architecture for text-to-image synthesis, showing a Mutli-Tailed Word-level Initial Generation stage, a Multi-Headed Spatial Dynamic Memory based refinement stage with three refinement heads, and image prediction.}
    \label{fig:msdmgan}
\end{figure*}

Our MSMT-GAN architecture (Figure \ref{fig:msdmgan}) comprises of three stages - a Multi-Tailed Word-level Initial Generation (MTWIG) stage, and two refinement stages. Each refinement stage is Multi-Headed, and each refinement head has a separate Spatial Dynamic Memory (SDM) module. Section \ref{subsection:DIG} presents our MTWIG stage, Section \ref{subsection:SDM} presents our SDM module for a single refinement head, and the details of our Iterative Multi-Headed Mechanism (IMHM) are presented in Section \ref{subsection:MIR}.
\subsection{Multi-Tailed Word-level Initial Generation (MTWIG)}
\label{subsection:DIG}
We highlight that previous multi-stage methods \citep{zhang2017stackgan, Zhang2018stackgan++, li2019controllable, zhu2019dm} all rely on the same type of initial generation stage and focus only on improving the refinement stages - making the conventional assumption that the performance of multi-stage generators is primarily determined by the refinement stages, and that the quality of the "rough initial image" is of little importance. In our paper, we break from this tradition and demonstrate for the first time that gains can be achieved in the final stage of image refinement by making an improvement to the initial images. \\\\
The conventional approach synthesizes initial images directly from a sentence-level vector without attempting to separate image attributes at a word-level. As a result, words that are entangled at the sentence-level representation generate initial image attributes that are inherently ambiguous in nature. In our novel Multi-Tailed Word-level Initial Generation (MTWIG) stage, we overcome this shortcoming by explicitly creating separate sets of image attributes for each word n-gram.\\\\
First, we sample a vector of random noise $z$ from a normal distribution and use a pretrained text-encoder to extract a sentence-level vector and word-level vectors: $s$ and $W$ from the input text.
\begin{gather}
    W = \{w_1, w_2, ..., w_L\}; \hspace*{0.1cm}w_l \in \mathbb{R}^{N_w} \;\hspace*{0.5cm};\hspace*{0.5cm}\; s \in \mathbb{R}^{N_s} \;\hspace*{0.5cm};\hspace*{0.5cm}\; z_n \sim \mathcal{N}(0,1); \hspace*{0.1cm}z \in \mathbb{R}^{N_z}
\end{gather}
Where $L$ is the number of words in the text-description, and $N_z$, $N_s$ and $N_w$ are the dimensions of the noise vector, sentence vector and word vectors respectively. To mitigate over-fitting, the Conditioning Augmentation technique \citep{zhang2017stackgan} is used to resample the sentence-vector from an independent Gaussian distribution. This resampled sentence vector $s'$ and the noise vector $z$ are then concatenated with each word-level vector $w_l$ from the input text sequence, and the sequence of concatenated vectors are passed through a 1D convolutional operation $V$ of stride 1 (see Figure \ref{fig:msdmgan}).
\begin{gather}
    F = V(\{concat(s', \;z,\; w_l) \;|\; \;\forall\; w_l \in W \})
\end{gather}
The length $T$ of the output sequence $F$ depends on the kernel size used by $V$ and the vectors of the output sequence $f_t \in F$ are each separately passed through a series of upsampling blocks to generate corresponding sets of image features $S_t$. These sets of image features or "tails" each correspond to a different word n-gram from the input text sequence. If we use a kernel size of 1 for $V$, then each tail $S_t$ corresponds to a single word. If we use a kernel size of 2, then each tail $S_t$ corresponds to a word bi-gram, and so on. We combine our sequence of tails $\{S_t\}$ together in an iterative fashion using the adaptive gating fusion mechanism introduced by \cite{zhu2019dm} (discussed in Section \ref{subsection:fuse}).
\begin{gather}
    S_{1:t} = fuse(S_{1:t-1},\;S_t,\; P^\text{MTWIG}, \rho^\text{MTWIG})\hspace*{0.5cm}; \hspace*{0.5cm}R_1 = S_{1:T}
\end{gather}
Where $P^\text{MTWIG}$ and $\rho^\text{MTWIG}$ denote parameter matrix and bias terms, $S_{1:t}$ denotes a combination of the first $t$ tails, and $S_{1:1}$ denotes the first tail $S_1$. The combination of all $T$ tails gives us the final image features $R_1$ for our initial stage. Notice that by concatenating each word vector $w_l$ with $s'$ and $z$ before the 1D convolution, each tail is created with some common information, so they may learn to fuse together coherently. Each upsampling block consists of a nearest neighbor upsampling layer and a 3×3 convolution operation. An initial image is predicted from $R_1$ using a 3×3 convolution. 
\subsection{Spatial Dynamic Memory (SDM)}
\label{subsection:SDM}
In this section, we describe the operation of a single refinement head. Unlike previous methods, our novel Spatial Dynamic Memory (SDM) module creates a separate region-contextualized text representations for each image region. This allows us to interpret the same text in fundamentally different ways at different parts of an image and assimilate region-contextualized information from text at each part. To begin with, we have the set of word-level vectors $W$ and image features $R_{k-1}$ from the previous stage of generation.
\begin{gather}
    {R_{k-1} = \{r_{1,1}\;,\; r_{1,2}\;, ...,\; r_{s,s}\}\;;\;\hspace*{0.1cm} r_{u,v} \in \mathbb{R}^{N_r}} 
\end{gather}
Where $|s \times s|$ is the number of image pixels and $N_r$ is the dimension of pixel features. We obtain refinement features in three steps: \textit{Memory Writing}, \textit{Key Addressing} and \textit{Value Reading}.  \\
\hspace*{0.3cm}\textbf{Memory Writing}: 
First, we divide the fine-grained $s \times s$ inital image into a coarse $h \times h$ sized grid-map and average the pixel features within each grid-cell to get grid-level image features $C$.
\begin{gather}
    C_{i,j} = \frac{1}{|p\times p|}\hspace*{0.2cm}\sum_{u=(i-1)*p +1}^{i*p}\hspace*{0.2cm}\sum_{v=(j-1)*p + 1}^{j*p}r_{u,v}
\end{gather}
Where $p = s/h$, so that $|p\times p|$ are the number of pixels represented by each grid cell. Then, we create $L \times h \times h$ memory slots $\{m_{l,i,j}\}$ - one corresponding to each word $l$ for each grid-cell $(i,j)$. These slots are our region-contextualized representations of each word, and each slot uses a separate memory writing gate $g_{l,i,j}^w$ to fuse information from each grid-cell $(i,j)$  with each word feature $w_l$. 
\begin{gather}
    g_{l,i,j}^w(R_{k-1}, w_L) = \sigma\left(A*w_l + B_{i,j}*C_{i,j}\right)\\
    m_{l,i,j} = M_w(w_l) \odot g_{l,i,j}^w + M_c(C)_{i,j} \odot (1 - g_{l,i,j}^w)
\end{gather}
The grid-level features $C$ are encoded using a 2d convolution operation $M_c$ (with stride 1 and $N_m$ output filters) and we use a common 1x1 convolution operation $M_w$ to encode all word vectors to a $N_m$ dimensional space. $A$ and $B_{i,j}$ are $1\times N_w$ and $1\times N_r$ matrices respectively.\\
\hspace*{0.3cm}\textbf{Key Addressing:} In this step, we compute attention weights $\{\alpha_{l,i,j,a,b}\}$ over our region-contextualized text-representations $\{m_{l,i,j}\}$. The dimensions $(a,b)$ index pixels within grid-cell $(i,j)$, so that each slot $m_{l,i,j}$ gets a matrix $\alpha_{l,i,j}: p \times p$ of attention weights. Each weight is computed as a similarity probability between a key-encoding of the slot: $\phi_K(m_l)_{ij}$ and a query vector: $q_{i,j,a,b}$, where $\phi_K(.)$ is a 2d convolution operation with stride 1 and ($N_{r} + p^2)$ output filters. 
\begin{gather}
    \alpha_{l,i,j,a,b} = \frac{\exp(\phi_K(m_l)_{i,j} *q_{i,j,a,b})}{\sum_{l=1}^{L} \exp(\phi_K(m_l)_{i,j}*q_{i,j,a,b})}
\end{gather}
In the case of single headed image refinement, we use the previous image features $R_{k-1}$ to obtain the query vectors. A query vector $q_{i,j,a,b}$ is made up of three components, 1)A global-level query: $q^{global}$, 2)A grid-level query: $q_{i,j}^{grid}$, and 3)A pixel-level query: $q_{i,j,a,b}^{pixel}$. To obtain these three components, we encode $R_{k-1}$ using three separate 2d convolution operations: $\phi_{Q^{global}}(.)$, $\phi_{Q^{grid}}(.)$ and $\phi_{Q^{pixel}}(.)$, each with a stride of 1 and $N_r$ output filters. 
\begin{gather}
\label{eq:query_formation}
Q^{global} = \phi_{Q^{global}}(R_{k-1})  \;\hspace*{0.2cm};\hspace*{0.2cm}\; Q^{grid}  = \phi_{Q^{grid}}(R_{k-1}) \;\hspace*{0.2cm};\hspace*{0.2cm}\; Q^{pixel} = \phi_{Q^{pixel}}(R_{k-1})
\end{gather}
Then, the average of all pixel features of $Q^{global}$ becomes the global-level query component $q^{global}$. The average of pixel features within the grid cell $(i,j)$ of $Q^{grid}$ becomes the grid-level query $q_{i,j}^{grid}$, and the pixel feature at location $(a,b)$ within grid cell $(i,j)$ is extracted from $Q^{pixel}$ to give us the pixel-level query component $q_{i,j,a,b}^{pixel}$. 
\begin{gather}
    q^{global} = \frac{1}{|s\times s|}\sum_{u=1}^{s}\sum_{v=1}^{s}Q^{global}_{u,v}\hspace*{0.5cm};\hspace*{0.5cm} q^{grid}_{i,j} = \frac{1}{|p\times p|}\hspace*{0.2cm}\sum_{u=(i-1)*p +1}^{i*p}\hspace*{0.2cm}\sum_{v=(j-1)*p + 1}^{j*p}Q^{grid}_{u,v}
    \\
    \label{eqn:pixelquery}
    q_{i,j,a,b}^{pixel} = Q^{local}_{h(i,a), \hspace*{0.1cm}h(j,b)} \;\hspace*{0.5cm};\hspace*{0.5cm}\;h(i,a) =(i-1)*p + a
\end{gather}
Where $(h(i,a),\;h(j,b))$ indexes the pixel at location $(a,b)$ within grid-cell $(i,j)$. To obtain the final query $q_{i,j,a,b}$, we concatene these three components together.\\
\hspace*{0.3cm}\textbf{Value Reading:} In the value reading step, for each pixel $(a,b)$ within a grid-cell $(i,j)$, we compute a weighted sum of value-encoded memory slots: $\phi_V(m_l)_{ij}$ along the word dimension $l$. 
\begin{gather}
    e_{i,j,a,b} = \sum_{l=1}^{L}\alpha_{l,i,j,a,b}\cdot\phi_V(m_l)_{i,j}
\end{gather}
$\phi_V(.)$ is a 2d convolution operation with stride 1 and $N_r$ output filters. We now have $e_{i,j}: p \times p \times N_r$ dimensional matrices - each of which corresponds to a single grid cell of our coarse $h \times h$ grid map. To obtain $s \times s$ fine-grained refinement features, we apply the mapping:  
\begin{gather}
    o_{h(i,a)\;, \;h(j,b)} = e_{i,j,a,b}  
\end{gather}
Where $h(.,\;.)$ is the function defined in Eq.\ref{eqn:pixelquery}. That is, we populate each grid cell with $|p \times p|$ vectors of $N_r$ dimensionality. Since $p=s/h$, we are left with a set of refinement features $O =\{o_{u,v}\}$ that are made up of $|s \times s|$ vectors of $N_r$ dimensionality, each of which corresponds to a single pixel.
\subsection{Iterative Multi-Headed Mechanism (IMHM)}
\label{subsection:MIR}
Current methods generate refinement features only once at each refinement stage and attempt to address all image aspects in a single-shot. This limits the precision with which each refinement stage can learn to improve the prior image. In order to make it easier for each refinement stage to precisely address multiple image aspects, we introduce a novel iterative multi-headed mechanism that makes multiple distinct modifications to the prior image features under common discriminator feedback. Each head of our mechanism has a separate spatial dynamic memory module formed from $R_{k-1}$ and $W$. For the first refinement head, we use the previous image features $R_{k-1}$ to obtain a query matrix and extract a set of refinement features $O_1$ exactly as described in Section \ref{subsection:SDM}. Then, we fuse $O_1$ and $R_{k-1}$ using the fusion mechanism introduced by \cite{zhu2019dm} (described in Section \ref{subsection:fuse}) to obtain an updated set of image features $U_1$. If we use only a single refinement head, then this becomes our response for the refinement stage $k$. However, if we use more than one refinement head, then for the next head, we use $U_1$ to obtain a query matrix. That is, we follow the same mechanism outlined in Section \ref{subsection:SDM}, but replace $R_{k-1}$ with $U_1$ in Eq.\ref{eq:query_formation}. Doing so, we extract a second set of refinement features $O_2$, and we fuse $O_2$ and $U_1$ to obtain updated image features $U_2$. We proceed in this iterative fashion until we have used all of our refinement heads. The final updated image features are fused with the original image features $R_{k-1}$ in a skip-connection to obtain the final response of the refinement stage $k$. That is, if we have $T$ refinement heads:
\begin{gather}
    U_t = fuse(U_{t-1},\; O_{t},\; P_t,\; \rho_t) \;\hspace*{0.5cm};\hspace*{0.5cm}\;response_k = fuse(U_{T},\; R_{k-1},\; P^{skip},\; \rho^{skip}) 
\end{gather}
Notice, we use separate parameter matrix and and bias terms $P$ and $\rho$ for each fusion operation, so that we combine refinement features and image features in a custom way for each head. The, $response_k$ is passed through several residual blocks \citep{he2016deep} and an upsampling block to obtain higher resolution image features $R_k$. Each block consists of a nearest neighbor upsampling layer and a 3×3 convolution operation. Finally, a refined image $x_k$ is predicted from $R_k$ using a 3×3 convolution operation.
\subsection{Objective Function}
\label{subsection:OBJfunc}
The objective function for our generator network is defined as: 
\begin{gather}
    L = \lambda_1 L_{CA} + \sum_{k=2}^K \lambda_2 L_{RED_k} + \sum_{k=1}^K \left(L_{G_k} + \lambda_3 L_{DAMSM_k}\right)
\end{gather}
Where, $L_{CA}$ denotes the conditioning augmentation loss \citep{zhang2017stackgan}, $G_{k}$ denotes the generator of the $k^{th}$ stage so that $L_{G_k}$ denotes the adversarial loss for $G_k$, $L_{RED_k}$ denotes our redundancy loss for the $k^{th}$ stage, and $L_{DAMSM_k}$ denotes the DAMSM text-image matching loss \citep{xu2018attngan} for the $k^{th}$ stage. $\lambda_1$, $\lambda_2$ and $\lambda_3$ are hyperparameters that combine the various losses.\\
\hspace*{0.3cm}\textbf{Redundancy Loss:}
To encourage each head of a refinement stage to focus on separate image aspects, we average region-wise information of each head's output refinement features and penalize similarity between different refinement heads. That is, for $T$ refinement heads: 
\begin{gather}
    f(t) = \frac{1}{|s\times s|}\sum_{u=1}^{s}\sum_{v=1}^{s}o_{u,v} \hspace*{0.1cm} \;\hspace*{0.5cm};\hspace*{0.5cm}\;
    L_{RED_k} = \sum_{i=1}^T\;\sum_{j=i+1}^T sim(f(i), f(j))
\end{gather}
Where  $o_{u,v} \in O_t \text{ in stage } k$ (see Section \ref{subsection:MIR}) and $sim$ is the cosine similarity between vectors. We call this sum of pairwise similarity $L_{RED_k}$ the ``redundancy loss" of the $k^{th}$ refinement stage.\\
\hspace*{0.3cm}\textbf{Adversarial Loss:} The adversarial loss for $G_k$ is defined as: 
\begin{gather}
\footnotesize
    L_{G_k} = -\frac{1}{2}[\mathbb{E}_{x\sim p_{G_k}} \log D_k(x) + \mathbb{E}_{x\sim p_{G_k}} \log D_k(x,s)]   
\end{gather}
$D_{k}$ is the discriminator network for the $k^{th}$ stage of generation. The first term provides feedback on image realism independent of the input text, and the second term provides feedback on the realism of the image in light of the input text. Alternate to adversarial training of $G_k$, each discriminator $D_k$ is trained to classify images as real or fake
by minimizing the discriminator loss $L_{D_k}$.
\begin{equation}
\tiny{
\begin{split}
    L_{D_k} = & \underbrace{-\frac{1}{2}[\mathbb{E}_{x\sim p_{data}} \log D_k(x) + \mathbb{E}_{x\sim p_{G_k}} \log (1 - D_k(x))]}_\text{unconditional loss} + \underbrace{-\frac{1}{2}[\mathbb{E}_{x\sim p_{data}} \log D_k(x,s) + \mathbb{E}_{x\sim p_{G_k}} \log (1 - D_k(x,s))]}_\text{conditional loss}
\end{split}}
\end{equation}
Where the unconditional component distinguishes synthetic and real images independent of the input text, and the conditional component distinguishes them in light of the input text.
\section{Experiments}
\label{section:Experiments-label}
\textbf{Datasets:} We evaluate our method on the Caltech-UCSD Birds 200 (CUB) dataset \citep{wah2011caltech} and the Microsoft Common Objects in Context (COCO) dataset \citep{lin2014microsoft}. The CUB dataset contains 8,855 training images and 2,933 test images, with 10 corresponding text
descriptions for each image. The COCO dataset, contains 82,783 training images and 40,504 test images, with 5 corresponding text descriptions for each image. We preprocess the datasets according to the methods introduced by \cite{zhang2017stackgan}.\\\\
\textbf{Evaluation metrics:} To evaluate the realism of images, we rely on the Fréchet Inception Distance (FID) \citep{heusel2017gans}. FID computes the distance between synthetic and real-world image distributions based on features extracted from a pre-trained Inception v3 network. A lower FID indicates greater image-realism. To evaluate the relevance of a synthetic image to it's generating text-description, we rely on the R-precision introduced by \cite{xu2018attngan}. R-precision is computed as the mean accuracy of using each synthetic image to retrieve one ground truth text-description from among 100 candidates. To evaluate a model on a dataset, we generate 30,000 synthetic images conditioned on text descriptions from the unseen test set.\\\\
\textbf{Implementation Details: }To obtain a sentence-level vector and word-level vectors for a given text description, we use the pretrained bidirectional LSTM text encoder employed by AttnGAN \citep{xu2018attngan}. Our MTWIG stage synthesizes images features with 64x64 resolution. Two refinement stages refine these features to 128x128 and 256x256 resolution respectively. At refinement stages, we use $T=6$ refinement heads and we use $h = 8$ to divide each input image into a coarse $8\times8$ grid map. We use the same discriminator network architecture employed by \cite{zhu2019dm}. Further implementation details are provided in Section \ref{subsection:AID}. 
\subsection{Ablative Experiments}
\begin{figure*}[!h]
    \centering
    \includegraphics[width=\linewidth]{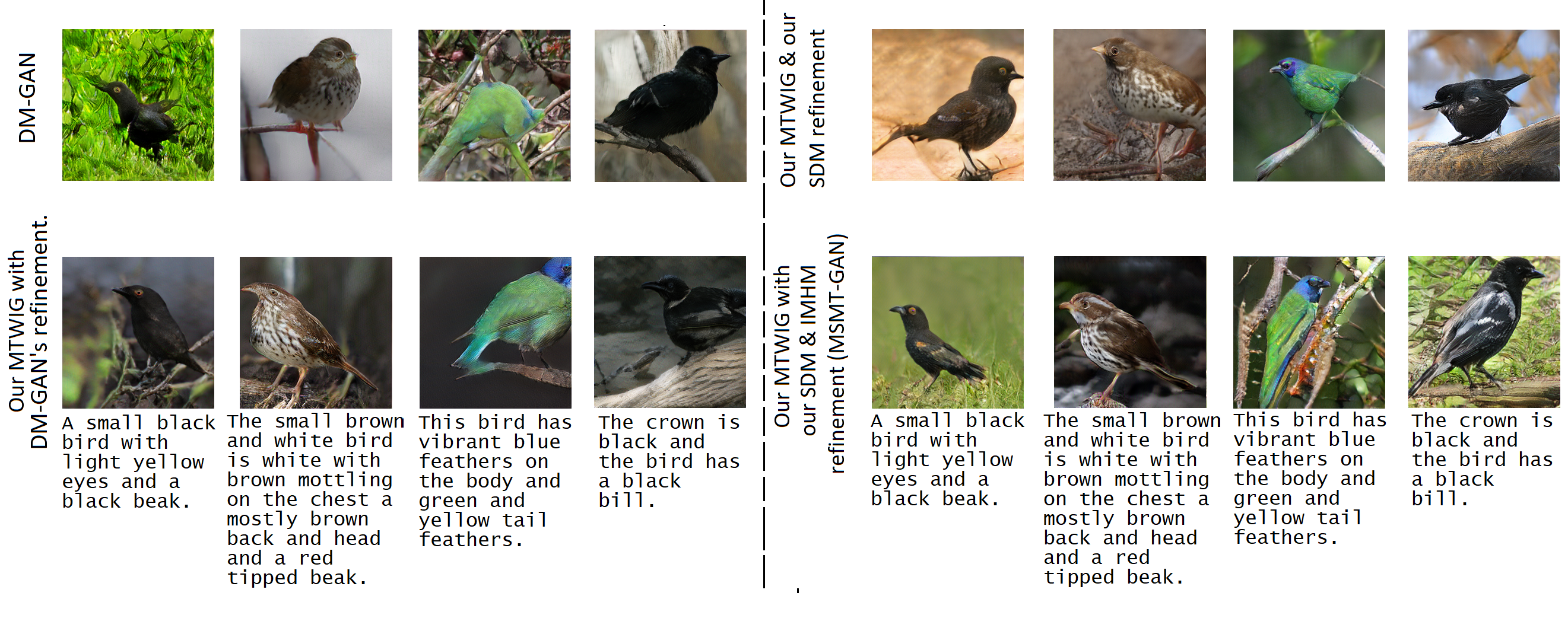}
    \caption{Comparison of DM-GAN with ablative versions of our model trained for 700 epochs on the CUB dataset.}
    \label{fig:ablation}
\end{figure*}

\textbf{Effectiveness of Multi-Tailed Word-level Initial Generation:}  In our experiments (Appendix A.1), we find that our MTWIG stage is most effective when used with a kernel size of 3, so that we generate a separate tail for each word tri-gram. To evaluate the effectiveness of our MTWIG(ks=3) stage in multi-stage models, we train our MTWIG(ks=3) method with DM-GAN \citep{zhu2019dm} style-refinement stages for 700 epochs on the CUB dataset, and observe that it decreases the FID score achieved by the original DM-GAN model by 7.72\% and increases R-precision by 2.76\% (Table \ref{table:t2}). Figure \ref{fig:ablation} shows the improved visual quality of the refined images. We again point out that previous multi-stage methods \citep{zhang2017stackgan, Zhang2018stackgan++, xu2018attngan, li2019controllable, zhu2019dm}) all rely on the same type of initial generation stage, and we expect a similar boost in performance if we replace their initial stage with our MTWIG stage. \\\\
\textbf{Effectiveness of Spatial Dynamic Memory:} In order to evaluate the effectiveness of our SDM based-refinement stages, we compare a multi-stage model that uses our MTWIG(ks=3) stage and DM-GAN's refinement stages against a multi-stage model that uses our MTWIG(ks=3) stage and our single-headed SDM-based refinement stages. Both models are trained on the CUB dataset for 700 epochs. We observe that our SDM-based refinement out-performs DM-GAN's refinement, decreasing FID score by 4.64 \% , and boosting R-precision by an additional 0.83\% (Table \ref{table:t2}). Figure \ref{fig:ablation} shows that SDM-based refinement generates images of better visual quality than those formed by DM-GAN's refinement stages for the same initial generation architecture. \\\\
\textbf{Effectiveness of the Iterative Multi-Headed Mechanism:} To evaluate the effectiveness of our IMHM refinement, we compare the model that uses MTWIG(ks=3) and single-headed SDM-based refinement stages against our full MSMT-GAN model - that uses MTWIG(ks=3) and six SDM-based refinement heads for each stage. As before, both models are trained for 700 epochs on the CUB dataset. We find that refinement stages that use our multi-headed mechanism out-perform single-headed refinement stages, decreasing FID score 2.38\%, and boosting R-precision by another 1.03\% (Table \ref{table:t2}). Visually, besides overall better image quality, we find that images generated by multi-headed refinement stages posess text-irrelevant content that is far more detailed than that observed in images generated by single-headed refinement stages (Figure \ref{fig:ablation}).  

\begin{table}[!h]
  \caption{FID and R-precision of DM-GAN and ablative versions of our model. (With all of our variants trained for 700 epochs on CUB)}
  \label{table:t2}
  \centering
  \begin{tabular}{lll}
    \toprule
    \multicolumn{2}{c}{\hspace*{6.5cm} CUB} \\
    \cmidrule(r){2-3} 
    Method & FID $\downarrow$    & R-prcn (\%) $\uparrow$      \\
    \midrule
    DM-GAN & 11.91 & 76.58 ± 0.53\\
    \midrule
    MTWIG w/ DM-GAN's refinement & 10.99	 & 79.37 ± 0.73   \\
    MTWIG w/ SDM refinement & 10.48 & 80.20 ± 0.67 \\
    MTWIG w/ SDM and IMHM refinement (MSMT-GAN) & \textbf{10.23} & \textbf{81.23 ± 0.68} \\
    \bottomrule
  \end{tabular}
\end{table}

\subsection{Comparison with State of the Art}
\label{subsection:Sota-label}
\begin{figure*}[!h]
    \centering
    \includegraphics[width=\linewidth]{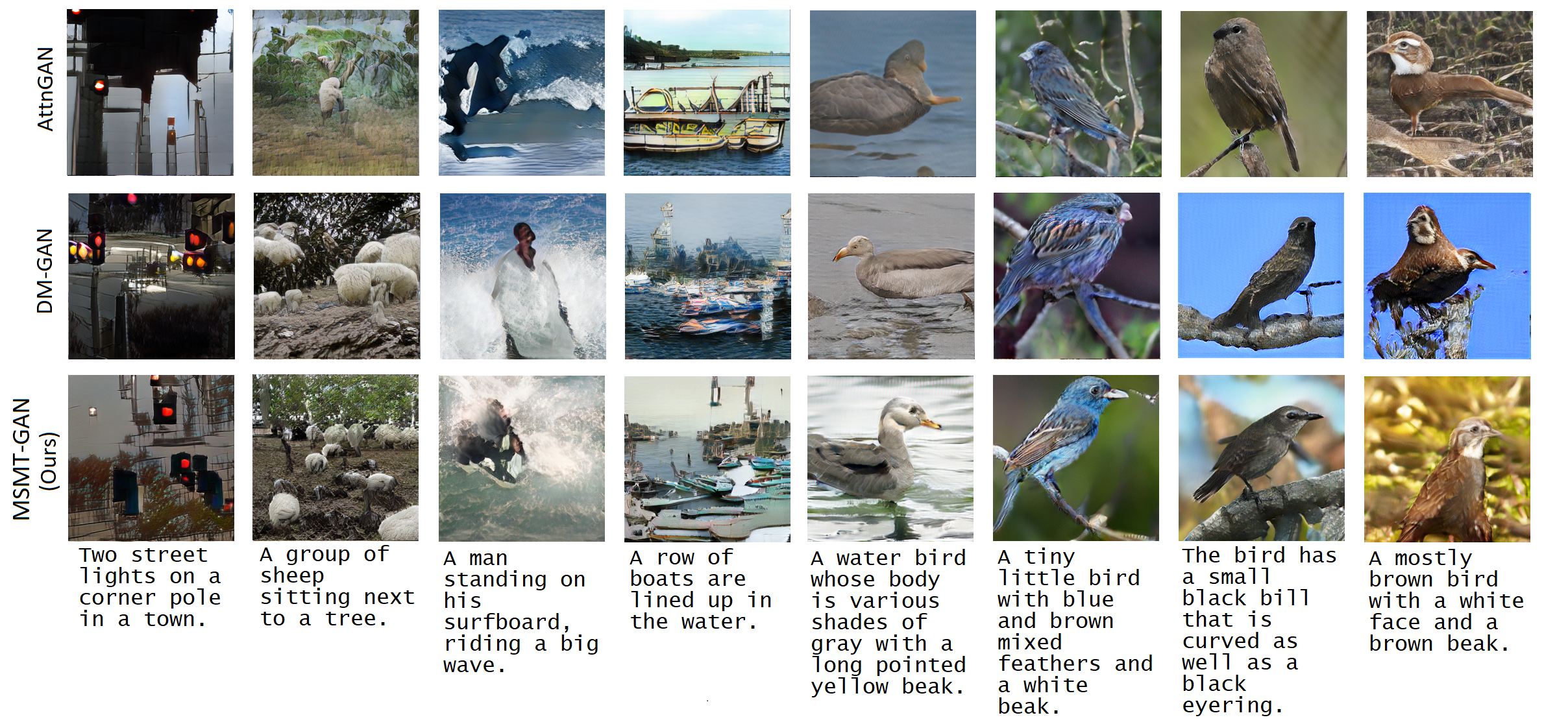}
    \caption{Comparison of MSMT-GAN with state-of-the art models on the CUB and COCO datasets.}
    \label{fig:sota}
\end{figure*}

To compare our architecture against the previous state-of-the-art for the task of text-to-image generation, we train MSMT-GAN models for 1000 epochs on the CUB dataset and for 210 epochs on the COCO dataset. As shown in Table \ref{table:t3}, our MSMT-GAN decreases the previous lowest reported FID score by 21.58\% on the CUB dataset -marking a significant improvement in image realism, and also boosts the previous best reported CUB R-precision by 4.24 \% - marking a large improvement in the similarity of synthetic images to their generating text. As shown in Table \ref{table:t4} and Table \ref{table:t3}, our model is comparable in size to previous methods, and outperforms the next closest contender of similar size for COCO (DM-GAN) by 4.21\% on FID score  -making it highly competitive with the current state-of-the-art. We also observe a slight improvement of 0.23\% on COCO R-precision. Qualitatively, we observe that synthetic images generated by our model are typically sharper and more realistic than those generated by prior methods (Figure \ref{fig:sota}). In particular, we observe that our method generates synthetic images that possess greater detail and that are better matches for the generating text. 
\begin{table}[!h]
  \caption{Number of parameters required at test-time (including text-encoders) for the previous state-of-the-art in comparison to our MSMT-GAN (approximate values reported in millions).}
  \label{table:t4}
  \centering
  \begin{tabular}{lll}
    \toprule
    Method & \# Paramers for CUB & \# Parameters for COCO\\
    \midrule
    AttnGAN & 9.16M & 22.43M\\
    ControlGAN  & 30.72M & 45.36M\\
    DM-GAN & 23.52M & 30.96M\\
    DF-GAN &  14.32M & 20.87M\\
    XMC-GAN & - &>111M\\
    \midrule
    Our MSMT-GAN & 48.11M & 55.16M\\
    \bottomrule
  \end{tabular}
\end{table}
\begin{table}[!h]
  \caption{FID and R-precision of the previous state-of-the-art and our MSMT-GAN trained for 1000 epochs on CUB and 210 epochs on COCO.}
  \label{table:t3}
  \centering
  \begin{tabular}{lllll}
    \toprule
    \multicolumn{2}{c}{\hspace*{4cm} CUB} & \multicolumn{2}{c}{\hspace*{4cm} COCO}    \\
    \cmidrule(r){2-3} \cmidrule(r){4-5}
    Method & FID $\downarrow$    & R-prcn (\%) $\uparrow$  & FID $\downarrow$    & R-prcn (\%) $\uparrow$\\
    \midrule
    AttnGAN\textsuperscript{\ref{note1}} & 14.01  & 67.82 ± 4.43 & 29.53 & 85.47 ± 3.69 \\
    ControlGAN & - & 69.33 ± 3.23 &  - & 82.43 ± 2.43  \\
    DM-GAN\textsuperscript{\ref{note1}}  & 11.91 & 76.58 ± 0.53 & 24.24 & 92.23 ± 0.37 \\
    DF-GAN\textsuperscript{\ref{note1}} & 13.48 & - & 33.29 & - \\
    \midrule
    Our MSMT-GAN& \textbf{9.34} & \textbf{80.82 ± 0.54} & \textbf{23.22} & \textbf{92.46 ± 0.28} \\
    \bottomrule
  \end{tabular}
\end{table}
\section{Conclusion}
In this work, we have proposed the MSMT-GAN architecture for the task of text-to-image synthesis. First, we introduced a novel Multi-Tailed Word Level Initial Generation stage (MTWIG) that explicitly generates separate sets of image features for each word n-gram. Second, we proposed a novel Spatial Dynamic Memory (SDM) module to contextualize text representations by image-region. Third, we  introduced a novel Iterative Multi-Headed Mechanism (IMHM) of image refinement to make it easier for each refinement stage to precisely address multiple image aspects. Our ablative experiments clearly demonstrate the effectiveness each of these three components, and we have shown that we are able to boost the performance of prior methods by replacing their initial stage with our MTWIG stage. Experiment results further demonstrate that our MSMT-GAN model significantly out-performs the previous state of the art on the CUB dataset, decreasing the lowest reported FID score by 21.58\% and boosting the CUB R-precision by 4.24\%. On the COCO dataset, we have demonstrated that MSMT-GAN is highly comptitive with current methods based on image realism and model size. In future work, we aim to design a discriminator model that provides more region-specific feedback than existing methods, to use in conjunction with our MSMT-GAN generator architecture.
\footnotetext[1]{\label{note1}We make our comparisons against the pretrained models released by the authors, and we report results using the official implementations of FID score.}
\section{Acknowledgements}
This work was performed while Amrit Diggavi Seshadri was a Post Baccalaureate Fellow at the Robert Bosch Center for Data Science and Artificial Intelligence (RBC-DSAI), at the Indian Institute of Technology Madras (IIT Madras). We would like to thank the center for sponsoring this fellowship and for providing us with sufficient resources.
\bibliography{arxiv}
\bibliographystyle{arxiv}

\appendix
\section{Appendix}
\subsection{Fusing two sets of image features together}
\label{subsection:fuse}
Given any two sets of image/refinement features $A=\{a_{u,v}\}$ and $B=\{b_{u,v}\}$ such that $a_{u,v},\;b_{u,v} \in \mathbb{R}^{N_r}$, we can use the adaptive gating mechanism introduced by \cite{zhu2019dm} to obtain a combined/updated set of features $\{r_{u,v}\}$, where $r_{u,v} \in \mathbb{R}^{N_r}$.
\begin{gather}
    g^r_{u,v} = \sigma(P* concat(a_{u,v}, \;b_{u,v}) +\rho)\\
    r_{u,v} = a_{u,v} \odot g^r_{u,v} + b_{u,v} \odot (1 - g^r_{u,v})
\end{gather}
Where $(u,v)$ index pixel locations, $N_r$ is the dimension of image pixel features, $g^r_{u,v}$ is a gate for information fusion, $\sigma$ is the sigmoid function and $P$ and $\rho$ are the parameter matrix and bias term respectively. In our paper, we invoke this adaptive gating mechanism by a function ``$fuse$" such that $fuse(A, B, P, \rho) = \{r_{u,v}\}$. 
\subsection{Analysis of Multi-Tailed Word-level Initial Generation}
\label{subsection:Comp}
\begin{table}[!h]
  \caption{Quantitative comparison of Fréchet Inception Distance and R-precision between previous Initial Generation (IG) and our Multi-Tailed Word-level Initial Generation (MTWIG) stage for varied kernel sizes (ks) on the CUB and COCO datasets.}
  \label{table:t1}
  \centering
  \begin{tabular}{lllll}
    \toprule
    \multicolumn{2}{c}{\hspace*{4cm} CUB} & \multicolumn{2}{c}{\hspace*{4cm} COCO}    \\
    \cmidrule(r){2-3} \cmidrule(r){4-5}
    Method & FID $\downarrow$    & R-prcn (\%) $\uparrow$     &  FID $\downarrow$    & R-prcn (\%) $\uparrow$    \\
    \midrule
    IG & 119.39 & \textbf{83.54 ± 0.57} & 51.22 & 94.57 ± 0.53  \\
    MTWIG (ks=1) & 118.49 & 82.78 ± 0.58 & 41.79 & 94.47 ± 0.49  \\
    MTWIG (ks=2) & 120.76 & 82.82 ± 0.92 & 40.23 & 94.65 ± 0.39 \\
    MTWIG (ks=3)  & \textbf{115.38} & 82.94 ± 0.74 & \textbf{39.85} & \textbf{94.80 ± 0.29}\\
    \bottomrule
  \end{tabular}
\end{table}
\begin{figure*}[!h]
    \centering
    \includegraphics[width=\linewidth]{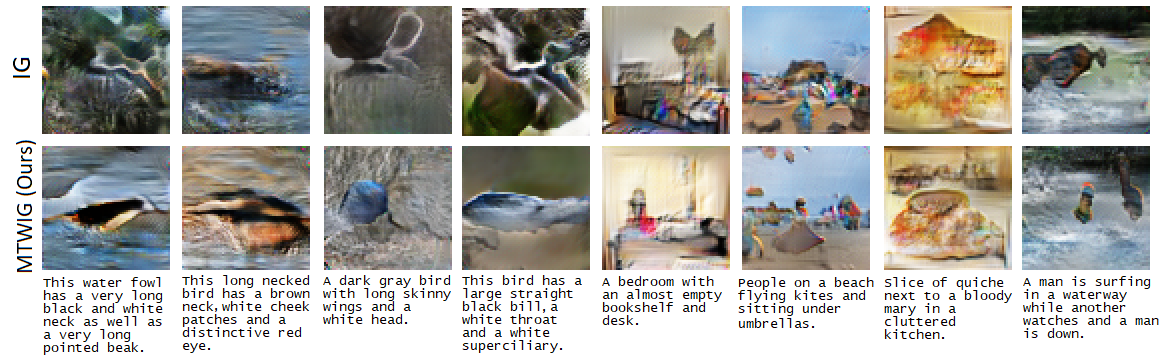}
    \caption{Multi-Tailed Word-level Initial Generation (MTWIG; kernel size=3) in comparison to conventional Initial Generation (IG).}
    \label{fig:disentangled}
\end{figure*}

To analyze the effectiveness of our MTWIG stage, we make comparisons with the inital generation (IG) stage employed by previous methods \citep{zhang2017stackgan, Zhang2018stackgan++, xu2018attngan, li2019controllable, zhu2019dm}. We train IG and MTWIG stages without refinement for 300 epochs on the CUB dataset and for 60 epochs on the COCO dataset. Table \ref{table:t1} shows a quantitative comparison between IG and MTWIG architectures that use different kernel sizes. We observe that our MTWIG method achieves best results for ks=3 (that is, by forming separate sets of word tri-grams), decreasing FID scores from the previous IG method by 3.36\% on the CUB dataset and by 22.2\% on the COCO dataset. The larger improvement observed on the COCO dataset highlights that our MTWIG method is most beneficial for complex scene generation, where the presence of a large number of distinct objects demands word-level separation of attributes. Our MTWIG stages also achieves competitive R-precision scores on both datasets, demonstrating that images synthesized by our method are well conditioned on their input text-descriptions. In Figure \ref{fig:disentangled}, we visually compare 64x64 images generated by IG and images generated by our MTWIG(ks=3) model. We observe that images generated by our method typically posses object attributes that are better separated and more clearly discernible.

\subsection{Additional Implementation Details}
\label{subsection:AID}
We set $N_w = 256$, $N_r = 48$ and $N_m = 96$ to be the dimension of text, image and memory feature vectors respectively. We set the hyperparameters $\{\lambda_1 = 1, \; \lambda_2 = 0.5, \; \lambda_3 = 5\}$ for the CUB dataset and $\{\lambda_1 = 1, \; \lambda_2 = 0.5, \;\lambda_3 = 50\}$ for the COCO dataset. All networks are trained using an ADAM optimizer \citep{kingma2015adam} with $\beta_1 = 0.5$ and $\beta_2 = 0.999$. The learning rate is set to be 0.0002 and we use a batch size of 20 for all networks. We train our models on a single Tesla V100 gpu, and observe that similar to previous multi-stage methods, training our MSMT-GAN method is a GPU intensive process. We require approximately 27min for one epoch of the CUB dataset, and approximately 4.7 hours for one epoch of the COCO dataset.
\subsection{Note on Evaluation Metrics}
We point out that there is inconsistency in the FID implementation used to evaluate prior methods. While some methods \citep{li2019object} report scores using the official Tensorflow version of FID - which uses the weights of a pretrained Tensorflow inception model, other methods \citep{zhu2019dm, zhang2021cross} have reported scores using an unofficial Pytorch implementation of FID - that uses the weights of a pretrained Pytorch inception model.\\\\ Each of these implementations computes different values of FID scores for the same sets of images, and scores computed from the two versions are \textbf{not} directly comparable with each other. We highlight that the correlation between FID score and human judgement is an empirical observation, that is dependent on the weights of the pretrained inception model used. The correlation with human judgement has \textbf{only} been shown to hold true for the weights of the pretrained Tensorflow inception model (used by the FID authors -\cite{heusel2017gans}), and has \textbf{not} been verified for the unofficial Pytorch inception model (which has different weights). As such, to ensure that we correlate with human judgement, in our paper we only report scores using the official Tensorflow implementation of FID score - computing values for prior work from the pretrained models released by their authors. In Table \ref{table:t5} below, we additionally report Pytorch-implementation FID scores of SEGAN \citep{tan2019semantics} and XMC-GAN \citep{zhang2021cross} models, as reported by their authors. It was not possible for us to recompute these scores using the official Tensorflow version of FID as pretrained models for these methods have not been made publicly available.\\\\ We again highlight, that the FID scores computed by the official Tensorflow implementation are not directly comparable with the scores computed by the unofficial Pytorch implementation.

In Table \ref{table:t5}, we additionally benchmark the performance of models on the Inception Score (IS) \citep{goodfellow2014generative}. We note however, that the FID score (which is reference-based and compares the distributions of real and synthetic images together) has been observed to be more consistent with human judgement of image realism than IS (which is reference-free and does not make comparisons to real images) \citep{heusel2017gans}. We were not able to recompute other metrics for RiFeGAN \citep{cheng2020rifegan} and LeciaGAN \citep{qiao2019learn} as the pretrained models have not been made publicly available.

\begin{table}[!h]
  \caption{FID and R-precision and IS of the previous methods and our MSMT-GAN trained for 1000 epochs on CUB and 210 epochs on COCO.}
  \label{table:t5}
  \centering
  \begin{tabular}{lllllll}
    \toprule
    \multicolumn{3}{c}{\hspace*{4cm} CUB} & \multicolumn{3}{c}{\hspace*{4cm} COCO}    \\
    \cmidrule(r){2-4} \cmidrule(r){5-7}
    Method & FID $\downarrow$    & R-prcn (\%) $\uparrow$     & IS $\uparrow$ & FID $\downarrow$    & R-prcn (\%) $\uparrow$     & IS $\uparrow$\\
    \midrule
    AttnGAN\textsuperscript{\ref{note1}} & 14.01  & 67.82 ± 4.43 & 4.36 ± 0.03 & 29.53 & 85.47 ± 3.69 & 25.89 ± 0.47     \\
    ControlGAN & - & 69.33 ± 3.23 & 4.58 ± 0.09 & - & 82.43 ± 2.43 & 24.06 ± 0.60  \\
    SEGAN & 18.16\textsuperscript{\ref{note2}}  & - & 4.67 ± 0.04 & 32.28 & - & 27.86 ± 0.31  \\ 
    RiFeGAN & - & - & \textbf{5.23 ± 0.09} & - & - & - \\ 
    LeciaGAN & - & - & 4.62 ± 0.06 & - & -& - \\
    DM-GAN\textsuperscript{\ref{note1}}  & 11.91 & 76.58 ± 0.53 & 4.71 ± 0.06 & 24.24 & 92.23 ± 0.37 & \textbf{32.43 ± 0.58} \\
    DF-GAN\textsuperscript{\ref{note1}} & 13.48 & - & 4.70 ± 0.06 & 33.29 & - & 18.64 ± 0.64\\
    XMC-GAN & - & - & - & \textbf{9.33}\textsuperscript{\ref{note2}} & - & 30.45  \\
    \midrule
    Our MSMT-GAN& \textbf{9.34} & \textbf{80.82 ± 0.54} & 4.55 ± 0.06 & 23.22 & \textbf{92.46 ± 0.28} & 28.91 ± 0.35\\
    \bottomrule
  \end{tabular}
\end{table}
In Table \ref{table:t5}, ``-” represents cases where the data was not reported or is reported in a manner which is non-comparable (besides FID values). 
\footnotetext[1]{\label{note1}We make our comparisons against the pretrained models released by the authors, and we report results using the official implementations of FID score.}
\footnotetext[3]{\label{note2} We use the FID score reported by the paper, but note that this was computed using an unofficial Pytorch implementation of FID - which is not directly comparable with the official FID implementation scores. See Section \ref{subsection:NOF} for further details.}
\label{subsection:NOF}
\subsection{More example images}
Figures \ref{fig:sota-extended-CUB} and \ref{fig:sota-extended-COCO} below show more images generated by our method in comparison to prior work on the CUB and COCO datasets. Again, we observe that synthetic images generated by MSMT-GAN are typically sharper and more realistic than those generated by previous methods. We also observe that the synthetic images generated by MSMT-GAN are usually a better match for the generating text.  
\begin{figure*}[!h]
    \centering
    \includegraphics[scale=0.3]{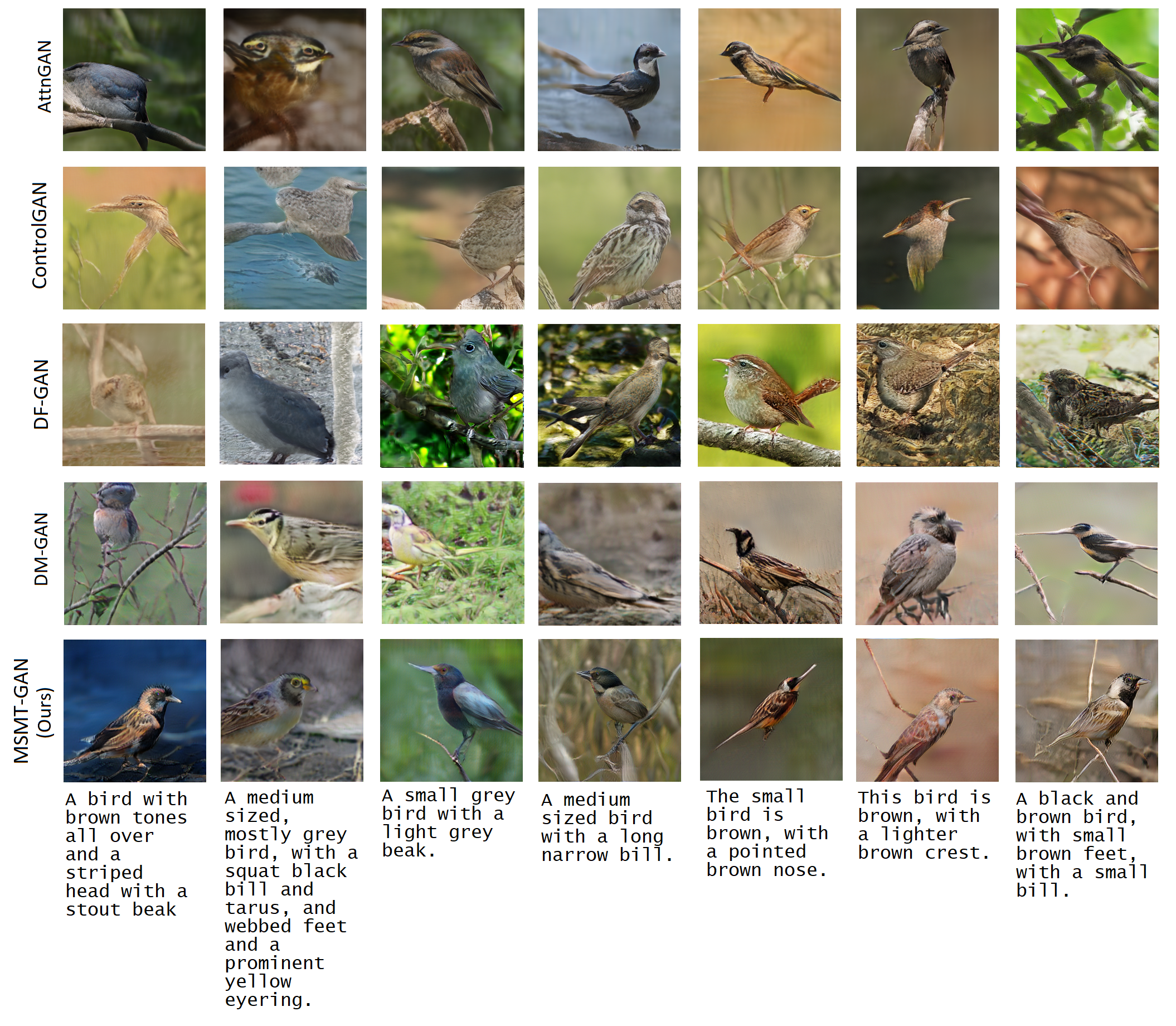}
    \caption{Extended comparison of MSMT-GAN with state-of-the art models on the CUB dataset.}
    \label{fig:sota-extended-CUB}
\end{figure*}
\begin{figure*}[!h]
    \centering
    \includegraphics[scale=0.3]{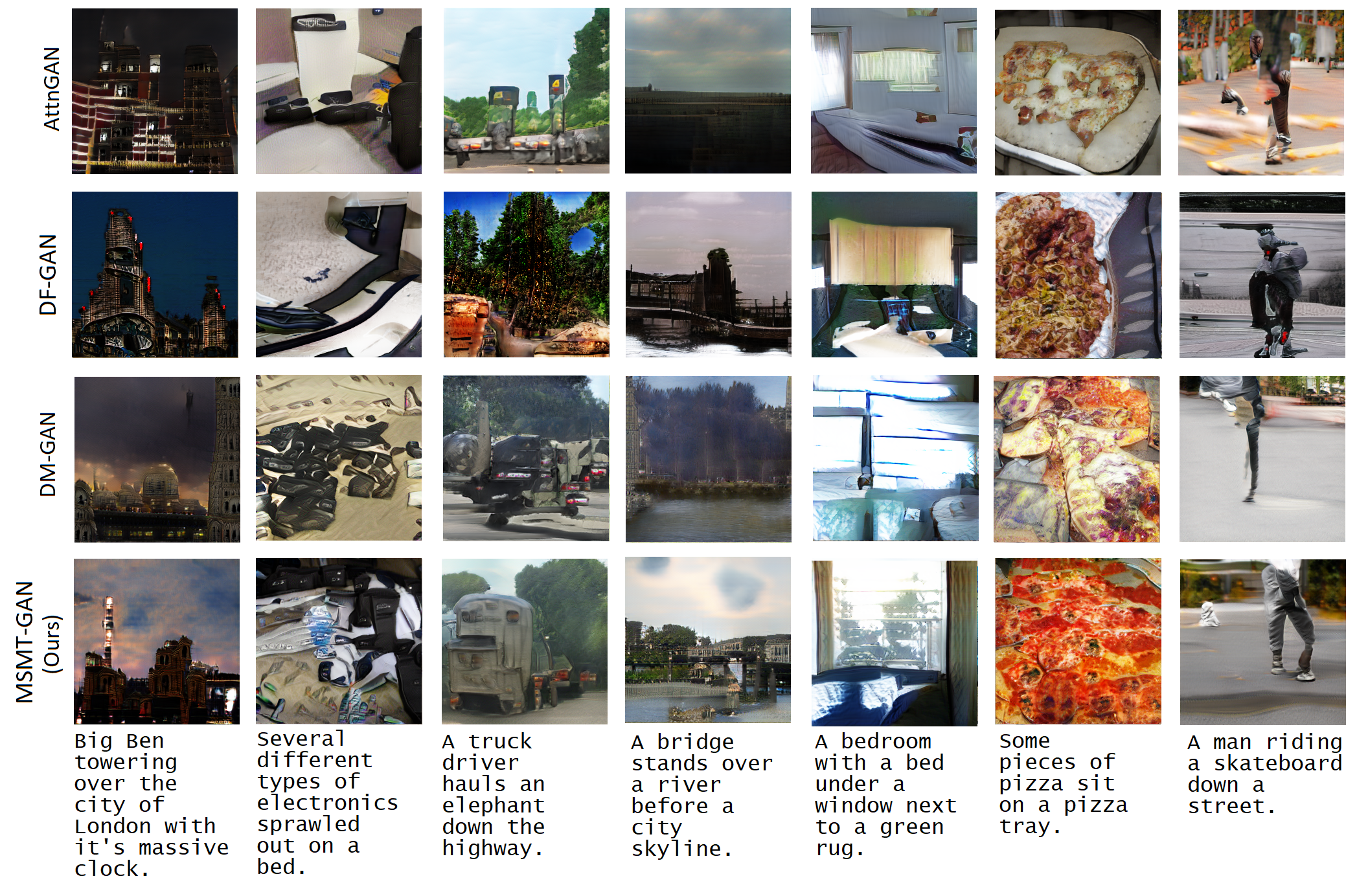}
    \caption{Extended comparison of MSMT-GAN with state-of-the art models on the COCO dataset.}
    \label{fig:sota-extended-COCO}
\end{figure*}
\end{document}